# Mammo-CLIP: Leveraging Contrastive Language-Image Pre-training (CLIP) for Enhanced Breast Cancer Diagnosis with Multi-view Mammography

Xuxin Chen[1], Yuheng Li[1,2], Mingzhe Hu[1,3], Ella Salari[1], Xiaoqian Chen[1],

Richard L.J. Qiu[1], Bin Zheng[4], Xiaofeng Yang[1,2,3*]

[1]Department of Radiation Oncology and Winship Cancer Institute, Emory University, Atlanta, GA 30322

[2]Department of Biomedical Engineering, Emory University and Georgia Institute of Technology, Atlanta, GA 30308

[3]Department of Biomedical Informatics, Emory University, Atlanta, GA, 30322

[4]School of Electrical and Computer Engineering, University of Oklahoma, Norman, OK 73019

*Correspondence: xiaofeng.yang@emory.edu

# ABSTRACT

Although fusion of information from multiple views of mammograms plays an important role to increase accuracy of breast cancer detection, developing multi-view mammograms-based computer-aided diagnosis (CAD) schemes still faces big challenges and no such CAD schemes have been used in clinical practice. To overcome these challenges, we investigate a new approach based on the concept of Contrastive Language-Image Pre-training (CLIP), which has sparked interest across various medical imaging tasks. By solving the challenges in (1) effectively adapting the single-view CLIP for multi-view feature fusion and (2) efficiently fine-tuning this parameter-dense model with limited samples and computational resources, we introduce a unique Mammo-CLIP, the first multi-modal framework to process multi-view mammograms and corresponding simple texts. Mammo-CLIP uses an early feature fusion strategy to learn multi-view relationships in four mammograms acquired from the craniocaudal (CC) and mediolateral oblique (MLO) views of the left and right breasts. To enhance learning efficiency, plug-and-play adapters are added into CLIP's image and text encoders for fine-tuning parameters and limiting updates to about 1% of the parameters. For framework evaluation, we assembled two datasets retrospectively. The first dataset, comprising 470 malignant and 479 benign cases, was used for few-shot fine-tuning and internal evaluation of the proposed Mammo-CLIP via 5-fold cross-validation. The second dataset, including 60 malignant and 294 benign cases, was used to test generalizability of Mammo-CLIP. Study results show that Mammo-CLIP outperforms the state-of-art cross-view transformer evaluated using areas under ROC curves (AUC= $0.841\pm0.017$ vs. $0.817\pm0.012$ and $0.837\pm0.034$ vs. $0.807\pm0.036$) on both datasets. It also surpasses previous two CLIP-based methods by 20.3% and 14.3% in AUC. Thus, this study highlights the potential of applying the finetuned vision-language models for developing next-generation, image-text-based CAD schemes of breast cancer.

# 1. Introduction

Breast cancer is the second leading cause of cancer death among women in the United States (Siegel et al., 2024). Mammography stands as the primary clinical imaging modality in population-based breast cancer screening programs and plays a crucial role in early breast cancer detection. Thanks to advancements in early detection techniques and therapeutic interventions over recent decades, breast cancer mortality rates have shown a notable decline. However, the interpretation of mammograms remains a challenging task for radiologists, as evidenced by the low diagnostic yield (e.g., detecting approximately 3.6 cancers per 1000 screening cases) (Kelly et al., 2010) and relatively high false positive recall rates (around 10%) (Nelson et al., 2016). Thus, to improve the accuracy of breast cancer detection and diagnosis as well as to reduce inter-reader variability among radiologists, computer-aided detection/diagnosis (CAD) schemes of mammograms have been developed and used in clinical practice for the last three decades aiming to offer radiologists a "second opinion" (Doi, 2007; Ko et al., 2006).

However, unlike radiologists who detect and diagnose suspicious breast lesions by reading 4 mammograms acquired from two views namely, craniocaudal (CC) and mediolateral oblique (MLO), of both left (L) and right (R) breasts simultaneously, the traditional CAD schemes detect suspicious breast lesions separately on single mammogram. Although the advent of deep convolutional neural networks (CNNs) (Chen et al., 2022a) provides a more powerful tool to develop CAD schemes than the traditional machine learning technologies using hand-crafted features or automatically computed radiomics features with generally higher performance (Danala et al., 2022; Dhungel et al., 2017; Yala et al., 2019), most existing CAD schemes using deep learning models are still single image based schemes trained solely on one-view mammogram. Such CAD schemes cannot fuse or integrate important and collected information depicting on multiple (4-view) mammograms into their decision-making processes. Additionally, current CAD schemes of mammograms also cannot utilize important domain-specific textual descriptions and categorizations, such as simple diagnostic descriptions of mammographic findings, anatomical annotations, or relevant patient history.

In the broader field of computer vision and natural language processing, recent advancements in Vision-Language Models (VLMs) (Chen et al., 2023; Zhang et al., 2024), pre-trained with both visual and linguistic data, have shown promising results in enhancing generalizability and interpretability, particularly in zero-shot and few-shot learning scenarios. Contrastive Language-Image Pre-training (CLIP) model (Radford et al., 2021) successfully integrates textual supervision into vision models. Hence, CLIP has garnered increasing interest in medical imaging for applications like zero-shot classification (Tiu et al., 2022), detection (Müller et al., 2022), and medical report generation (Huh et al., 2023; Xu et al., 2023). However, CLIP has not been extensively explored in developing CAD schemes of mammograms, particularly, the multi-view mammogram-based CAD schemes, to detect or classify breast lesions (Zhao et al., 2023).

The challenges in adapting these VLMs to develop multi-view mammogram-based CAD schemes are at least twofold. Firstly, CLIP is originally designed for and pre-trained on single-view images, making their adaptation to mammogram interpretation difficult. A standard mammography screening obtains 4 mammograms for each patient, namely LCC, RCC, LMLO and RMLO. Radiologists analyze these images simultaneously, with a focus on two crucial aspects of domain-specific multi-view information: bilateral asymmetry (Donnelly et al., 2024; Tan et al., 2016) and ipsilateral correspondence (Liu et al., 2021). Bilateral asymmetry refers to noticeable differences in the density, structure, or presence of lesions between the left and right breasts, as observed in mammograms. Bilateral asymmetry indicates either normal anatomical variations or signal potential pathological conditions, such as the developed lesions or other breast irregularities. In contrast, ipsilateral correspondence involves comparing findings within the same breast across different mammographic views. This comparison is critical for accurately confirming the presence and exacting location of abnormalities, ensuring that a lesion or other anomaly noted in one view is consistently identified in another view of the same breast, which can help reduce false-positive detection due to the impact of fibro-glandular tissue overlap in the different view of 2-D projection images (CC or MLO view mammograms). Secondly, directly finetuning CLIP using a traditional approach on a small-scale mammography dataset is not parameter-efficient, risking significant overfitting. For example, a popular publicly available mammogram dataset (DDSM (Heath et al., 2007; Lee et al., 2017)) that has been widely used by many researchers to develop CAD schemes includes multi-view mammograms of approximately 2,600 study cases. Its substantially smaller size compared to CLIP's pretraining data (400 million) can result in inferior transfer learning performance when fine-tuning CLIP using traditional methods. Thus, to develop highly performed CAD schemes of mammograms, other large datasets continue to be developed and assembled in the last several decades. A recently reported database (EMBED (Jeong et al., 2023)) includes both 2-D digital mammograms and 3-D digital breast tomosynthesis (DBT) images acquired from 116,000 women in which 20% of the dataset was made publicly available. However, developing large image databases with accurate annotations is very difficult, labor intensive and expensive. In addition, direct fine-tuning or pre-training from scratch on these large datasets pose challenges in computational resources.

To better address and/or overcome above two challenges, we propose a new multi-view mammogram analysis framework called *Mammo-CLIP* that enables us to (1) extend the capabilities of the traditional single-view image-based CLIP for integrating multi-view mammogram images into the learning process, and (2) achieve superior transfer learning performance using small mammogram datasets in a parameter-efficient manner. This approach not only differs from the late or middle-stage multi-view feature fusion methods used by existing CAD schemes but also emphasizes Mammo-CLIP's multimodal strengths, combining image and text data to enhance diagnostic accuracy. We summarize our contributions as follows.

- Early-stage multi-view feature fusion: We propose a novel strategy of early-stage feature fusion within CLIP's vision encoder to overcome its single-view image limitation. Specifically, the image encoder's transformer blocks are into local blocks for individual view processing and global blocks for multi-view feature fusion. The ratio of global to local blocks determines the stage at which feature fusion occurs - a higher proportion of global blocks facilitates earlier fusion, resulting in better classification performance.
- Parameter-efficient fine-tuning: Unlike traditional adapter-based strategies, we integrate adapters directly within each transformer block of both vision and text encoders to facilitate deeper co-adaptation of both image and textual features specific to multi-view mammogram analysis. This method proves more effective than positioning and fine-tuning a single adapter positioned externally to the encoders.
- VLM-based CAD scheme for multi-view mammogram analysis: Mammo-CLIP stands as the first VLM-based CAD framework specifically developed for multi-view mammogram analysis. This multi-modal framework utilizes simple yet effective mammography domain-specific texts to enhance predicting cased-based malignancy likelihood.

Furthermore, we recognize that the field of deep learning and artificial intelligence (AI) is continuously evolving, and more powerful VLMs will be released by various technology firms and institutions in the future. Our proposed framework is not dependent on a specific VLM, and it can readily adapt to new VLMs as they become available. This adaptability is particularly beneficial in the context of mammography, where the ability to assimilate domain-specific multi-view information is crucial. By aligning with new VLMs, our framework can leverage higher transfer learning efficiency and cost-effectiveness, ensuring that it remains up-to-date and effective in analyzing mammograms as AI technology advances. The objective of this study is to optimize and test the proposed Mammo-CLIP framework. The detailed methods and experimental results are reported in the following sections.

## 2. Related Work

**2.1 Multi-view CNNs for mammogram analysis**

By recognizing the limitation of the traditional single image-based CAD schemes of mammograms, deep learning-based CAD schemes of multi-view mammograms have gained significant research interest over the past decade. For instance, Carneiro et al. (2017) conducted multi-view mammogram analysis using ImageNet-pretrained CNNs. Specifically, the CNNs were finetuned to distinguish between malignant and benign microcalcification clusters and soft tissue masses, showing that models trained on both CC and MLO view mammograms substantially outperformed those trained on a single view mammogram. A later study (Wu et al., 2019) investigated more effective ways to fuse image features extracted from mammograms acquired from different views and breast sides. The researchers introduced a "view-wise" feature merging strategy using two tailored ResNet-22 models: one for processing two CC images and another for two MLO images of left and right breasts. This method produced separate predictions for CC and MLO images, which were then averaged during inference to make a final diagnostic decision. This approach was different from typical methods that fine-tune off-domain, natural image dataset-pretrained models, like those from ImageNet, for mammogram analysis. Instead, the authors began with in-domain pretraining on screening Breast Imaging Reporting and Data System (BI-RADS) classification, followed by fine-tuning for malignant/non-malignant and benign/non-benign classifications at the breast level. While the model demonstrated strong classification performance, the pretraining phase was lengthy, requiring about two weeks of training on four Nvidia V100 GPUs.

The methods discussed above fall within the track of implicitly merging the bilateral asymmetry and ipsilateral correspondence for multi-view mammogram analysis. Another track explicitly extracts and utilizes domain knowledge, considering the inherent geometric constraints of cross-view mammograms in building models for breast cancer detection and diagnosis. This usually requires preprocessing steps like matching ipsilateral breasts and registering bilateral breasts. Liu et al. (2020) presented a mammogram mass detection method using manually defined pseudo landmarks for aligning CC and MLO view mammograms and mapping geometric and semantic relations through a bipartite graph node mapping method. While effective in identifying the ipsilateral correspondence, the method lacks integration of the bilateral domain knowledge of two breasts, making it a two-view-one-side model. Addressing this gap, Liu et al. (2021) introduced an inception graph convolutional network (IGN) for bilateral view analysis, enabling four-view image reasoning to enhance breast mass detection. More recently, Jones et al. (2023) developed a two-view-two-side CAD framework using deep transfer learning and radiomics feature analysis for breast lesion classification. The approach needs bilateral image registration and matched region of interest (ROI) extraction from two bilateral mammograms.

Despite the efforts of the aforementioned approaches to explicitly mirror the process of radiologists in mammogram interpretation, significant challenges persist, particularly in accurately aligning locations between CC and MLO views of mammograms and achieving precise bilateral breast image registration. These difficulties are compounded by varying angles and difference in breast compression during imaging procedures, and possibly differences in two bilateral breast sizes. Such complexities impede the seamless application of pre-trained models in an end-to-end framework.

**2.2 Multi-view Transformers for mammogram analysis**

Vision Transformers (ViTs) (Dosovitskiy et al., 2020) have emerged as a strong alternative to CNN-based architectures for multi-view mammogram analysis. Central to ViTs is the self-attention mechanism, a key feature enabling these models to dynamically focus on the most relevant image areas or features pertinent to the task at hand. This self-attention mechanism is particularly effective in capturing long-range dependencies within the input sequence, a critical aspect of mammography. In recent studies of mammogram analysis, researchers have explored developing hybrid models that combine Transformers with CNNs. One study (Tulder et al., 2021) employed global cross-view Transformer blocks to transfer intermediate feature maps generated by the ResNet-18 model for both CC and MLO view mammograms. This technique primarily focuses on the fusion of ipsilateral features from two view (CC and MLO) mammograms of the same breast, though it does not explore bilateral mammographic feature asymmetry. Other studies (Wang et al., 2023) have adopted cross-view attention to merge bilateral-view representations from mammograms of both breasts. This approach yielded better classification results in a simplified task: separating cases with BI-RADS = 1 from BI-RADS > 1. Diverging from the hybrid approach, Chen et al. (2022b) demonstrated the effectiveness of a small, pure ViT in integrating information from all four mammogram views.

Another interesting direction in recent research integrates eye-gaze data with multi-view images to enhance mammogram classification performance (Ji et al., 2023). Utilizing a radiologist's eye-gaze tracking data as a supplementary annotation provides initial lesion localizations, enhancing the model's ability to detect abnormalities. Despite the progress in multi-view information fusion, the current methodologies are still confined to processing solely image-based data. They fall short in integrating both textual and visual data within mammography analysis.

**2.3 CLIP in multi-view image analysis**

The development of vision intelligence models has historically been confined to image data. Recent advances, particularly in large language models (LLMs) and VLMs, have shifted this paradigm by integrating text-based supervision. The substantial impact of text supervision in vision models was markedly amplified following the release of CLIP by OpenAI in early 2021 (Radford et al., 2021). CLIP relies on a self-supervised contrastive learning approach and extends its capabilities by training it with a dataset of 400 million image-text pairs sourced from the internet. CLIP's exposure to such a vast and varied dataset has enabled it to generalize effectively across a wide range of downstream tasks, including image or ROI detection, segmentation, and classification, etc. Its ability to perform zero-shot learning and classify images into unseen categories through textual descriptions makes it especially valuable in scenarios where large training data is unavailable. Recently, CLIP has emerged as a foundational model and has catalyzed interest in various medical subfields, such as pathology (Huang et al., 2023) and radiology (Zhao et al., 2023). In radiology, for instance, CLIP has been preliminarily applied to disease diagnosis tasks in chest X-rays (Tiu et al., 2022), organ segmentation, and tumor detection in CT scans (Liu et al., 2023), etc. Despite these advancements, the application of CLIP been primarily limited to single-view image analysis. Notably, there is a lack of research adapting CLIP for breast cancer diagnosis in mammography, where multi-view image data like screening mammograms are essential. Effectively leveraging complementary information across multiple view images requires specialized techniques. Consequently, this gap has restricted the broader application of CLIP in this domain.

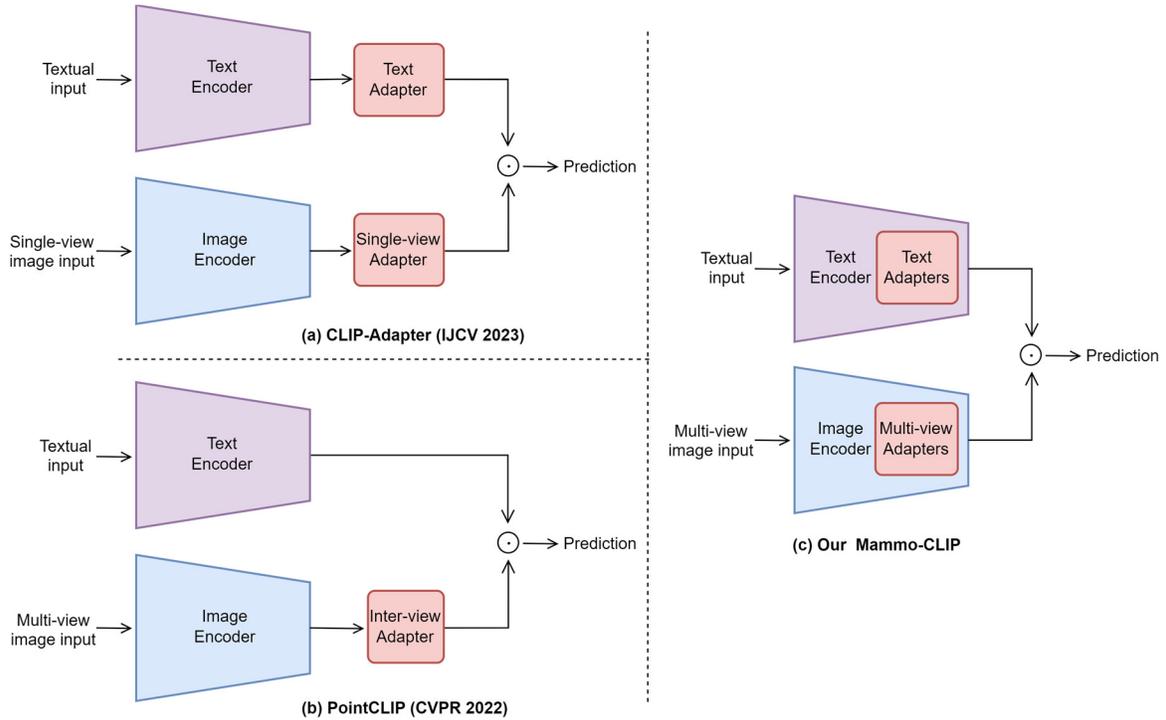

**Fig. 1.** Comparison of Adapter integration methods for multi-view image analysis between existing CLIP-based approaches and our proposed Mammo-CLIP. (a) CLIP-Adapter (Gao et al., 2023) inserts an image adapter and a text adapter outside the image and text encoders, respectively, but cannot process multi-view images. (b) PointCLIP (Zhang et al., 2022) inserts an inter-view adapter outside the image encoder for fusing multi-view information but does not adapt the text encoder. (c) Our Mammo-CLIP inserts multiple adapters inside both the image and text encoders, enabling better image-text co-adaptation for multi-view mammogram analysis.

In natural image analysis, we identified certain CLIP-based methods that can efficiently adapt the pretrained CLIP model for the downstream task of mammogram analysis. These methods like CLIP-Adapter (Gao et al., 2023) and PointCLIP (Zhang et al., 2022) employ adapters to realize parameter-efficient transfer learning on small datasets, making them promising candidates for extension to multi-view image analysis tasks with certain modifications. For example, CLIP-Adapter (Gao et al., 2023) adds two learnable bottleneck linear layers (adapters) (Houlsby et al., 2019) after both the image encoder and text encoder of the pretrained CLIP model, while keeping the CLIP backbone frozen during few-shot fine-tuning, as illustrated in Fig. 1a. The researchers explored three variants: 1) fine-tuning only the image branch adapter, 2) fine-tuning only the text branch adapter, and 3) fine-tuning both adapters, and found that combining both adapters did not improve performance over using only the visual (image) adapter, suggesting that the text and visual adapters might capture redundant or conflicting information. However, in our study, this observation did not hold true when inserting the text and visual adapters inside the encoders rather than appending them at the end. It is also worth noting that CLIP-Adapter, while initially designed for single-view natural images, can be adapted to process multi-view mammograms by concatenating feature representations from each view and feeding the combined input into the image encoder.

In contrast, PointCLIP (Zhang et al., 2022) (Fig. 1b), enables feature interaction for multi-view images by introducing a learnable inter-view adapter outside the visual branch. During fine-tuning, PointCLIP freezes CLIP's encoders and only fine-tunes the adapter. However, we found in our study that applying this late fusion to our multi-view mammogram analysis task yielded unsatisfactory performance. To explain this issue, we theorized that the late-stage feature fusion outside the CLIP model is disadvantageous, as the earlier layers are forced to process each view independently without accessing complementary inter-view information at different feature abstraction levels. In other words, this late fusion approach, by only considering high-level or decision-level representations, fails to leverage richer multi-view features from earlier stages, potentially limiting the model's ability to capture intricate multi-view interactions.

Considering the limitations of the aforementioned approaches, our study introduces the first image-text CAD framework for multi-view mammogram analysis using CLIP (Fig. 1c), aiming to outperform both traditional image-only CAD schemes and existing CLIP-based approaches. Should this novel framework prove successful, it could provide radiologists with an enhanced CAD-generated malignancy assessment score to differentiate between malignant and benign cases more accurately, thereby reducing false positives and unnecessary biopsies.

## 3. Materials and Methods

### 3.1 Image datasets

**Table 1**. Comparison of patient age and mammographic density distribution rated by radiologists using BI-RADS guidelines across two study datasets: internal 5-fold CV and external validation.

|  | Subgroup | Dataset 1: Internal 5-Fold CV | | Dataset 2: External Validation | |
|---|---|---|---|---|---|
|  |  | Negative Cases | Positive Cases | Negative Cases | Positive Cases |
| BI-RADS Density | 1 | 21 (4.4%) | 24 (5.1%) | 24 (8.2%) | 5 (8.3%) |
|  | 2 | 150 (31.3%) | 172 (36.6%) | 104 (35.4%) | 19 (31.7%) |
|  | 3 | 291 (60.8%) | 265 (56.4%) | 151 (51.4%) | 35 (58.3%) |
|  | 4 | 17 (3.5%) | 9 (1.9%) | 15 (5.1%) | 1 (1.7%) |
|  | Total | 479 | 470 | 294 | 60 |
| Patient Age | A < 40 | 30 (6.3%) | 24 (5.1%) | 7 (2.4%) | 0 |
|  | 40 ≤ A < 50 | 171 (35.7%) | 84 (17.9%) | 140 (47.6%) | 10 (16.7%) |
|  | 50 ≤ A < 59 | 160 (33.4%) | 113 (24.0%) | 74 (25.2%) | 18 (30.0%) |
|  | 60 ≤ A < 69 | 86 (18.0%) | 114 (24.3%) | 42 (14.3) | 17 (28.3%) |
|  | 70 ≤ A | 32 (6.7%) | 135 (28.7%) | 31 (10.5%) | 15 (25.0%) |
|  | Total | 479 | 470 | 294 | 60 |

We retrospectively assembled two image datasets from the existing de-identified full-field digital mammography (FFDM) image databases (Heidari et al., 2019; Zheng et al., 2012). The first dataset (Dataset 1) is a training dataset used for the model's few-shot fine-tuning through 5-fold cross-validation (CV), leading to internal evaluation. The second dataset (Dataset 2) is an independent testing dataset served for external validation to assess the model's generalizability and performance. Overall, these two datasets are collectively utilized for the development and evaluation of the diagnostic model. Table 1 summarizes and compares the patient age and mammographic density distributions rated by radiologists following BI-RADS guidelines across the two datasets. The first dataset comprises 3,796 mammograms acquired from 949 patients, spanning a wide age range, and including diverse breast density categories according to BI-RADS guidelines. Each case has four mammograms (LCC, RCC, LMLO and RMLO). Among these study cases, 349 were diagnosed as screening negative (no recall during screening) or benign by radiologists. The remaining cases were recommended for biopsy due to the detection of suspicious soft-tissue masses. Histopathological analysis of the biopsies confirmed 130 cases as benign and 470 as malignant. For binary classification using pre-trained CLIP models, we grouped both screening negative and benign lesion cases into a single cancer-free (negative) class, while confirmed malignant cases were categorized into a malignant or cancer (positive) class. This categorization supports our focus on predicting the likelihood of malignancy in mammographic cases. Consequently, this dataset comprises 479 negative cases and 470 positive cases, each verified through histopathological examination following the biopsy of lesions detected during mammographic screenings. Fig. 2 shows examples of a negative and a malignant case in the four standard views: LCC, RCC, LMLO, and RMLO.

Dataset 2 has 1,416 mammograms acquired from 354 women. The dataset includes 294 negative cases and 60 positive cases, which provides a more realistic representation of clinical mammography screening scenarios, where the majority of women are healthy or cancer-free, reflecting the low prevalence of malignancy typically observed in population-based screening dynamics.

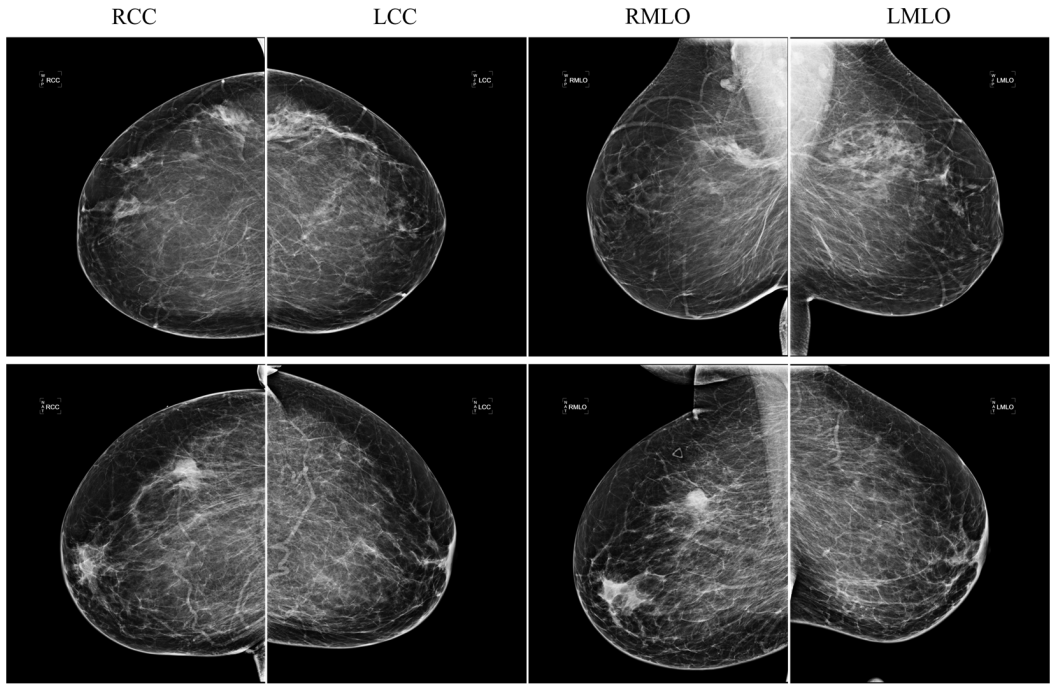

**Fig. 2.** Examples of breast cancer screening exams. First row (negative case): both breasts without any malignant findings; second row (positive case): right breast with a malignant finding and left breast without findings.

All mammograms were captured with Hologic Selenia digital mammography machines with a consistent pixel size of 70μm. To accommodate variations in breast size, images were acquired at two resolutions: 2,558×3,327 or 3,327×4,091pixels. Each mammogram is a 12-bit image with 4,096 grayscale levels. More detailed information of our FFDM image databases have been reported in our previous studies (i.e., (Zheng et al., 2012)). Following traditional CAD scheme protocols (Zheng et al., 2012), we applied a pixel averaging method with a 5×5 pixel kernel to subsample the original image resolutions to 512×666 and 666×819 pixels, with an increased pixel size of 0.35mm. To use pre-trained CLIP models for finetuning, the single-channel mammogram images were replicated and expanded into three-channel (RGB) images, and their pixel values normalized to the [0, 255] range. In addition, all images were resized to 224×224 pixels to match the standard input size of pre-trained models.

### 3.2 Parameter-efficient transfer learning with CLIP

CLIP learns transferable visual representations using natural language supervision. The architecture consists of two main components: a vision encoder and a text encoder. The vision encoder, implemented as a ResNet-50 (He et al., 2016) or a ViT (Dosovitskiy et al., 2020), extracts feature representations from single-view image inputs. The text encoder, typically built upon a transformer-based model like BERT (Devlin et al., 2018), generates text embeddings from input prompts. During pre-training, CLIP jointly optimizes the image and text encoders to learn a shared multi-modal embedding space. The objective is to maximize the cosine similarity between embeddings of matched image-text pairs while minimizing the similarity for mismatched pairings. By optimizing a symmetric cross-entropy loss over these similarity scores, this contrastive learning approach helps CLIP learn semantic associations between visual and textual information.

Due to CLIP's high number of parameters and the scarcity of training examples, directly fine-tuning the pre-trained model using conventional methods can lead to overfitting on the internal data and a significant performance drop on external validation sets. This highlights the challenging nature of transferring the learned knowledge of CLIP to downstream tasks via fine-tuning on few-shot examples. To address this issue, parameter-efficient transfer learning (PETL) (Houlsby et al., 2019), a technique originating from natural language processing, offers a solution by inserting lightweight adaptation modules, such as adapters, into pre-trained backbones. In the few-shot fine-tuning scenarios, PETL updates only the parameters of these adaptation modules while keeping the pre-trained weights frozen. By leveraging the pre-trained weights and adapting only a small set of parameters, PETL minimizes overfitting and achieves strong transfer learning performance, even with limited training samples.

**3.3 CLIP-driven framework for multi-view mammogram analysis**

Unlike the late-stage fusion approach that operates externally to the CLIP-based models or adapts only the image encoder, Mammo-CLIP performs early-stage feature fusion for multi-view mammograms within the CLIP architecture, while simultaneously co-adapting both the image and text encoders. This early fusion strategy, coupled with the joint adaptation of encoders, exposes the model to complementary cross-view features from the initial layers, enabling it to effectively capture and fuse multi-view feature relationships across various levels. As illustrated in Fig. 1c, our method inserts learnable adapter modules within CLIP's vision encoder and text encoder. These adapters dynamically adapt and fuse multi-view mammogram features at different levels during few-shot transfer learning. In contrast to CLIP-Adapter's ineffective text adapter fine-tuning (Gao et al., 2023) and PointCLIP' lack of a text adapter (Zhang et al., 2022), our approach co-finetunes both the visual and text adapters, resulting in superior performance.

Fig. 3 illustrates the proposed Mammo-CLIP framework for multi-view mammogram analysis. For fine-tuning Mammo-CLIP with Dataset 1 (Fig. 3a), we employ a supervised approach. Due to the absence of pre-existing textual descriptions for the mammographic cases, we generate two simple text descriptions based on their labels, each corresponding to one diagnostic category (positive or negative). On the other hand, we intentionally refrain from incorporating detailed diagnostic data such as patient histories or lesion characteristics. This ensures that all models operate on the same level of relevant information for malignancy prediction and prevents our multi-modal model from gaining an unfair advantage by utilizing textual information that single-modal models cannot effectively leverage in their decision-making processes. Instead, our text descriptions include only basic mammography domain-specific terminology, such as mammographic views and breast laterality, and uses "normal" for negative cases and "abnormal" for positive ones. These simple label-related texts and the corresponding 4-view mammograms are then fed into Mammo-CLIP's text and image encoders, respectively. During fine-tuning, the pre-trained encoder weights remain fixed, while only the adapter parameters are updated. The text encoder generates embeddings $T^n$ and $T^p$ for the negative and positive case label descriptions, and the image encoder performs early-stage multi-view feature fusion for the 4-view mammograms and outputs an image embedding $I$ for the case. Cosine similarity scores $[I.T^n, I.T^p]$ are calculated between the image embedding and the positive and negative text embeddings. These scores are passed through the SoftMax function to convert logits into a probability vector, which represents the model's predicted likelihood of the case being positive or negative. The cross-entropy loss is then calculated using the predicted probabilities and the true labels, guiding Mammo-CLIP to learn and improve its ability to predict a case's malignancy likelihood.

The fine-tuned Mammo-CLIP is used for zero-shot breast cancer classification on Dataset 2 (Fig. 3b). For each case with an unknown label, the model processes its 4-view mammograms to generate an image embedding. This embedding is compared to the embeddings of positive and negative text prompts. The classification is determined by the similarity scores between the case's image embedding and the text prompts, with the higher score indicating the predicted label. The following sections provide further details on the image encoder and text encoder of the proposed Mammo-CLIP framework.

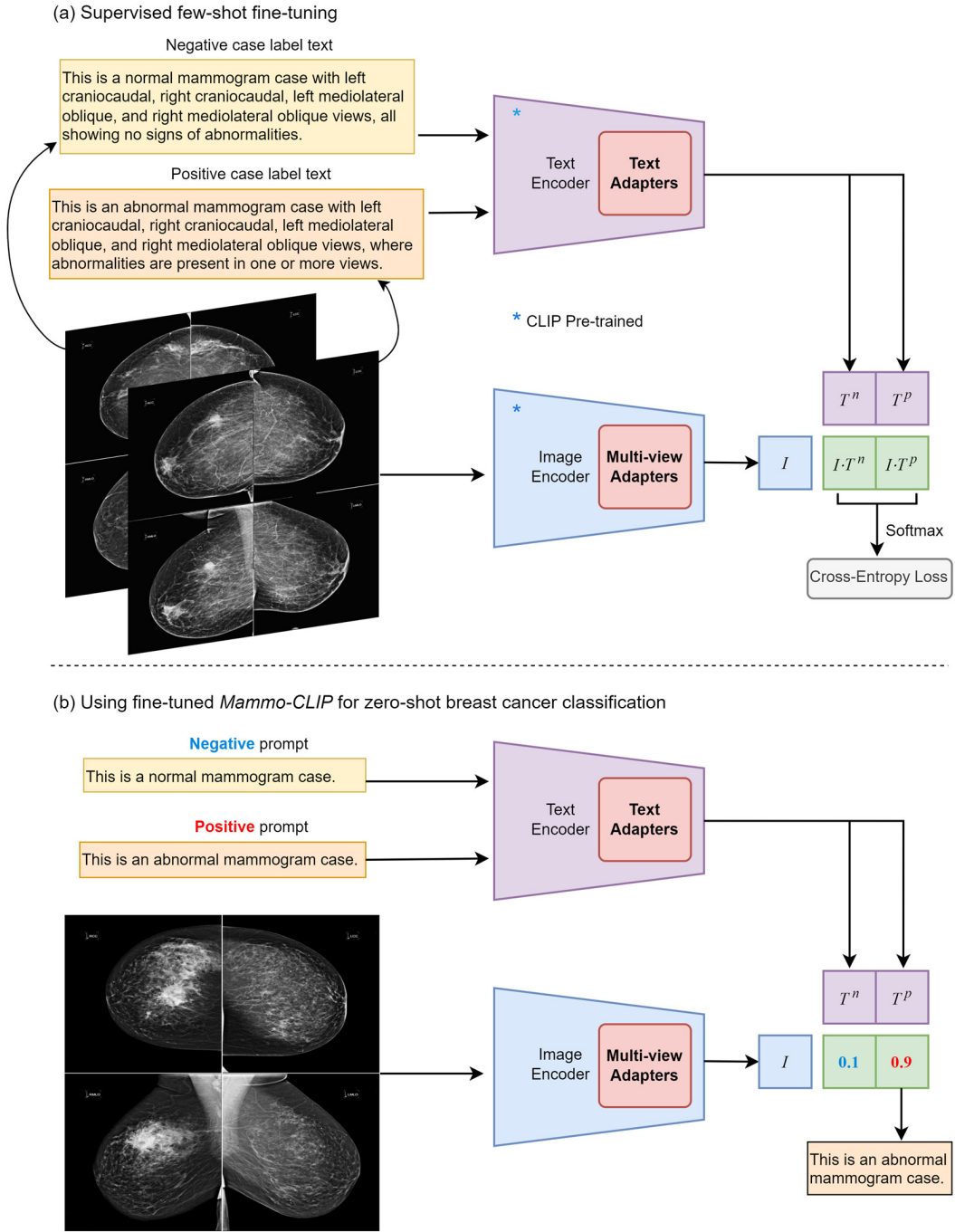

**Fig. 3.** Overview of the proposed Mammo-CLIP framework for multi-view mammogram analysis. (a) Supervised few-shot fine-tuning. Mammo-CLIP co-adapts the CLIP-pretrained image and text encoders using generated domain-specific text descriptions and 4-view mammograms, learning to predict malignancy likelihood. (b) Zero-shot breast cancer classification. The fine-tuned Mammo-CLIP classifies unseen cases by comparing the cosine similarity between their 4-view mammogram image embeddings and positive/negative prompt embeddings. The case is assigned the label corresponding to the text prompt that yields the higher similarity score with its image embedding.

### 3.3.1 Mammo-CLIP image encoder

#### Local Transformer blocks

Given a mammogram image $x \in \mathcal{R}^{H \times W \times C}$, it is projected into a sequence of image tokens $z \in \mathcal{R}^{L \times D}$, by a convolutional layer, where $L$ is the number of image tokens, $D$ is the embedding dimension of each token, $(H, W)$ and $C$ denote the image resolution and number of channels. A learnable classification token of dimension $D$, is prepended to the sequence of tokens. Thus, the mammogram image is converted to a token sequence $z \in \mathcal{R}^{(1+L) \times D}$. With position embedding, the token sequence serves as the input to the CLIP image encoder.

Since each patient has four mammograms acquired in one screening examination, the corresponding input sequence is denoted as $\{z_{LCC}, z_{RCC}, z_{LMLO}, z_{RMLO}\}$. We divide the image encoder's transformer blocks into $N_l$ local and $N_g$ global blocks. Local blocks independently learn patch relationships within each mammogram, while global blocks jointly process information from all four view images to capture inter-view dependencies for malignancy prediction. In this study, we employed three pre-trained CLIP image encoder backbones: ViT-B/32, ViT-B/16, and ViT-L/14. For ViT-B/32 and ViT-B/16, $N_l + N_g = 12$, and for ViT-L/14, $N_l + N_g = 24$. The sequence tokens from each mammography view pass through the local blocks sequentially, as shown in equation (1), and illustrated in Fig. 4.

$$z^0 = \{z^0_{LCC}, z^0_{RCC}, z^0_{LMLO}, z^0_{RMLO}\} = \{z_{LCC}, z_{RCC}, z_{LMLO}, z_{RMLO}\},$$

$$\tilde{z}^n = Adap^n_1(\text{MSA}(\text{LN}(z^{n-1})) + z^{n-1}), \quad n = 1 \ldots N_l \quad (1)$$

$$z^n = Adap^n_2(\text{MLP}(\text{LN}(\tilde{z}^n)) + \tilde{z}^n), \quad n = 1 \ldots N_l$$

where $LN$ represents layer normalization, MSA is multi-head self-attention, MLP is multi-layer perceptron, and $Adap^n_1$ and $Adap^n_2$ are the adapters for the $n^{th}$ MSA and MLP layers, respectively. The output of the current Transformer block serves as the input to the next Transformer block. For each of the four mammograms, the final local transformer block outputs a sequence of image features, as shown in equation (1). As shown in Fig. 4, the local blocks for four mammogram views share weights. This weight-sharing scheme offers one important benefit in reducing the tunable parameters introduced by the adapters. This approach is favored over using independent local block series for each mammogram, which reduces computational complexity and parameter count.

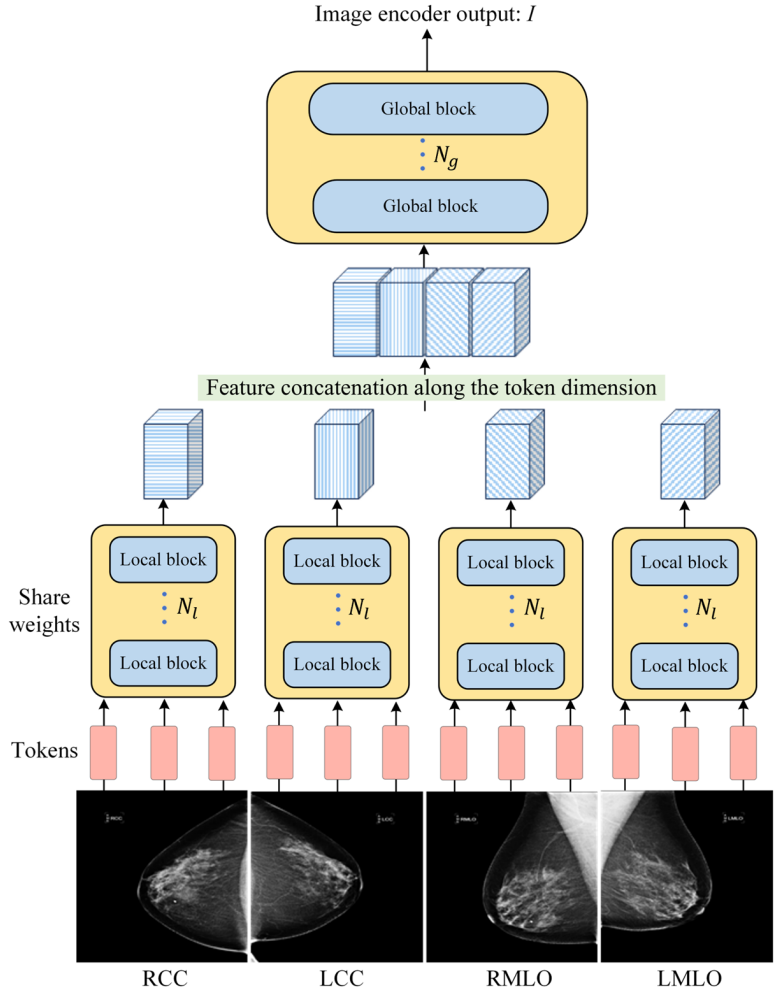

**Fig. 4.** Multi-view feature fusion using local and global transformer blocks within the image encoder.

*Global Transformer blocks*

The final local transformer block produces four distinct feature sequences for the four mammograms: $\{z_{LCC}^{N_l}, z_{RCC}^{N_l}, z_{LMLO}^{N_l}, z_{RMLO}^{N_l}\} \in \mathcal{R}^{(1+L) \times D}$. These feature sequences are subsequently concatenated along the token dimension. The concatenated sequence, represented by $g^0 \in \mathcal{R}^{(4+4L) \times D}$, is then passed through the global transformer blocks sequentially. Each global transformer block, like its local counterpart, comprises an original layer and MLP layer, along with their corresponding adapters, as depicted in equation (2).

$$\begin{aligned}\tilde{g}^n &= Adap_1^n(\text{MSA}(\text{LN}(g^{n-1})) + g^{n-1}), & n = 1 \dots N_g \\ g^n &= Adap_2^n(\text{MLP}(\text{LN}(\tilde{g}^n)) + \tilde{g}^n), & n = 1 \dots N_g\end{aligned} \quad (2)$$

By harnessing the transformer's strength to capture long-distance feature relationships, the global transformer blocks simultaneously process information from all four mammograms. This enables the implicit recognition and utilization of domain-specific knowledge pertaining to multi-view mammograms, such as bilateral asymmetry and ipsilateral correspondence. These characteristics cannot be obtained by analyzing a single view in isolation but are expected to enhance the classification results of breast cancer. The number of global transformer blocks relative to local transformer blocks determines the stage at which multi-view mammographic feature fusion occurs. A higher number of global transformer blocks and a correspondingly lower number of local transformer blocks result in earlier feature fusion. We will examine and discuss the impact of performing multi-view feature fusion at different stages inside the Mammo-CLIP's image encoder in later sections.

The class token from the final global block undergoes a linear projection head to map the image features from the image-specific embedding space to a shared multi-modal embedding space. Within this unified space, cosine similarity directly assesses the proximity between image and text embeddings. The scaled similarity scores across all image-text batch pairs are sent into the SoftMax function to convert logits into a probability vector. In terms of loss function, Mammo-CLIP slightly deviates from the original CLIP's symmetric cross-entropy loss by calculating the cross-entropy loss solely from image logits. This simplified loss enhances the image encoder's robustness for multi-view mammogram analysis by focusing primarily on image categorization and text matching without the need for reciprocal text-to-image alignment.

*3.3.2 Mammo-CLIP text encoder*

The text encoder of CLIP is a Transformer (Vaswani et al., 2017) with architecture modifications described in (Devlin et al., 2018). Like the image encoder, adapters are inserted after the MSA and MLP layers for each transformer block of the text encoder. We use two text descriptions to differentiate between the negative and positive classes. For fine-tuning Mammo-CLIP on Dataset 1 (Fig. 3a), the negative label text states, "This is a normal mammogram case with left craniocaudal, right craniocaudal, left mediolateral oblique, and right mediolateral oblique views, all showing no signs of abnormalities." In contrast, the positive label text reads, "This is an abnormal mammogram case with left craniocaudal, right craniocaudal, left mediolateral oblique, and right mediolateral oblique views, where abnormalities are present in one or more views." The text description is at the case level without going into image-level or breast-level details. The text description for each patient is first tokenized and then passed through the transformer blocks of the text encoder. Finally, the textual features are projected from the text-specific space to the shared multi-modal embedding space through a linear projection head.

For external validation on Dataset 2 (Fig. 3b), concise negative and positive prompt templates, such as "This is a normal mammogram case." and "This is an abnormal mammogram case.", suffice to achieve satisfactory performance.

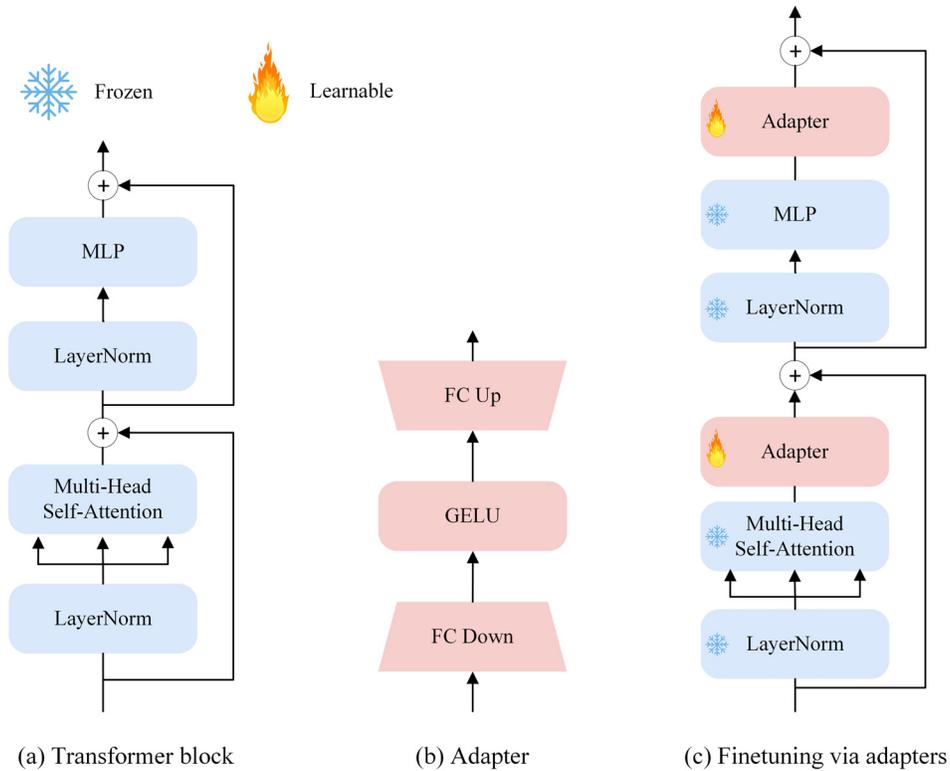

**Fig. 5.** Illustration of (a) a standard transformer block, (b) an adapter module, and (c) adapter finetuning.

The visual and text encoders of CLIP are composed of several transformer blocks, each containing a multi-head self-attention (MSA) layer followed by a multilayer perceptron (MLP) layer (Fig. 5a). The Adapter module (Fig. 5b) follows a simple yet effective bottleneck design (Houlsby et al., 2019), where an activation function, such as GELU, is positioned between two fully connected (FC) layers. The initial FC layer reduces the input feature dimensions by a predetermined bottleneck ratio (e.g., 16, 32, 64), while the subsequent FC layer reconstructs the features back to their original dimensionality. The bottleneck ratio is a hyperparameter controlling the number of trainable parameters during the fine-tuning process. A larger bottleneck ratio means less tunable parameters during few-shot finetuning, which is particularly important when working with small mammogram datasets. For simplicity, we adopt the default bottleneck ratio of 32 from previous research (Chen et al., 2024) to balance model complexity and performance. There are several possible configurations for incorporating adapters into the transformer blocks. Adapters can be inserted exclusively in the MSA layer, the MLP layer, or concurrently in both layers. Empirical evidence suggests that the simultaneous integration of Adapters in both the MSA and MLP layers generally enhances the model's adaptability. This is the strategy employed in this study. Furthermore, Adapters can be arranged either sequentially (Fig. 5c) or in parallel with the MSA and MLP components. Following the findings of a recent research (Chen et al., 2024), we opted to position the Adapters after these two layers in our implementation.

### 3.4 Evaluation of Mammo-CLIP performance

In this study, we investigated and compared the performance of Mammo-CLIP using three ViT backbones for the image encoder: ViT-B/32, ViT-B/16 and ViT-L/14. When implementing each Mammo-CLIP variant, we used Dataset 1 as a few-shot fine-tuning dataset. To obtain the optimal parameters and minimize the risk of overfitting, a 5-fold cross validation method was applied to build the final model. For each validation fold, the best-performing model was selected by considering the epoch that yielded the highest classification accuracy. Then, the finetuned Mammo-CLIP was tested using Dataset 2 that serves as an independent testing dataset. To provide a comprehensive comparison, we also finetuned and tested four state-of-the-art CAD models reported recently in the literature using the same two datasets, which include two CNN-based models (Image-wise CNN (Carneiro et al., 2017)) and View-wise CNN (Wu et al., 2019)) and two Transformer-based models (Cross-view transformer (Tulder et al., 2021) and Multi-view DeiT (Chen et al., 2022b)). Additionally, we compared Mammo-CLIP's performance with three CLIP-based models: CLIP (Radford et al., 2021), CLIP-Adapter (Gao et al., 2023), and PointCLIP (Zhang et al., 2022). Although these CLIP-based models were not originally developed for mammography but rather for natural images, we assessed and compared their potential for adaptation to the mammography domain. Therefore, Mammo-CLIP was compared against seven benchmark models.

To evaluate the few-shot classification performance of Mammo-CLIP on the balanced Dataset 1, we compute classification accuracy and areas under receiver operating characteristic (ROC) curves (AUC) based on the model-generated classification scores. To measure the zero-shot classification performance on the imbalanced Dataset 2, which has a higher proportion of negative cases, we replaced accuracy with metrics more sensitive to class imbalance. Specifically, we visualize the precision-recall curve (PRC) for each model using their probability estimates of malignancy and computed the area under the PRC (PRAUC), as the PRC has been shown more informative than the ROC curve when evaluating performance on datasets with a low ratio of positive (malignant) mammograms (Stadnick et al., 2021). AUC and PRAUC capture different facets of a predictive model's performance. The ROC curve reflects the trade-off between the true positive rate and the false positive rate, while the PRC summarizes the trade-off between precision (positive predictive value) and recall (true positive rate), both using different probability thresholds for classification (Wu et al., 2019). We use a publicly available ROC curve fitting program, ROCKIT (http://metzroc.uchicago.edu/MetzROC), to generate smooth ROC curves based on the maximum likelihood estimates of the model-generated prediction scores. The computed ROC and PRC curves, and evaluation indices are then compared to find out which image encoder backbone can help Mammo-CLIP yield the highest performance and whether Mammo-CLIP can achieve overall higher performance than the benchmark models.

### 3.5 Implementation details

For the three ViT backbones used for the image encoder (ViT-B/32, ViT-B/16, and ViT-L/14), they differ in the model size and patch size, influencing their ability to capture fine-grained details and complex features from mammograms. When fine-tuning Mammo-CLIP variants, we employ a parameter-efficient approach by updating only the adapters' weights while keeping the pre-trained weights of the image and text encoders frozen. Among the three CLIP-based comparison methods, PointCLIP (Zhang et al., 2022) and CLIP-Adapter (Gao et al., 2023) adopt the adapter-based fine-tuning method just like Mammo-CLIP, whereas CLIP (Radford et al., 2021), which does not have adapters, was fully fine-tuned with the pretrained weights of both the image and text encoders updated. Likewise, the rest benchmark models including CNN-based models (Carneiro et al., 2017), View-wise CNN (Wu et al., 2019), Cross-view transformer (Tulder et al., 2021) and Multi-view DeiT (Chen et al., 2022b)) were also fully fine-tuned.

To ensure fair model comparisons, (1) we largely followed the training setting in Touvron et al. (2021) and maintained consistency across all experiments. The models were fine-tuned for 400 epochs using the AdamW optimizer (Loshchilov and Hutter, 2018) with a batch size of 8 and a base learning rate of 5e-4. The learning rate was warmed up during the first 10 epochs and then decayed following a cosine schedule. (2) Standard data augmentations were applied uniformly across all experiments, including normalization, random horizontal flips, and random erasing. (3) All models were finetuned using only the entire mammography images with their corresponding case-level cancer labels, without resorting to any additional preprocessing steps such as pectoral muscle removal or background cropping. This guarantees that any performance differences can be attributed to the models themselves rather than variations in data preprocessing. All experiments were conducted using PyTorch on a single NVIDIA Tesla V100 GPU with 32GB of VRAM.

### 4. Results

### 4.1 Few-shot classification performance on Dataset 1

Table 2 summarizes the few-shot fine-tuning classification performance of 10 CAD models on Dataset 1. The result shows that two multimodal Mammo-CLIP models implemented with ViT-B/16 and ViT-L/14 backbones achieve higher accuracy and AUC than other existing image-based models ($p < 0.05$). For instance, Mammo-CLIP (ViT-L/14) demonstrates superior performance with an AUC of 0.836±0.019 and accuracy of 0.823±0.015, outperforming earlier CLIP-based models like CLIP (Radford et al., 2021), PointCLIP (Zhang et al., 2022), and CLIP-Adapter (Gao et al., 2023). These models also merge multi-view mammographic features within an image-text framework but do not reach the same level of effective feature fusion. The multimodal Mammo-CLIP also outperforms traditional image-based methods, including the convolution-based models, such as the view-wise CNN (Wu et al., 2019) (AUC = 0.807 ± 0.016, $p = 0.012$), and hybrid transformers like the cross-view transformer (Tulder et al., 2021) (AUC = 0.817 ± 0.012, $p = 0.033$). These methods typically fuse multi-view mammogram features at the middle or late stages.

Table 2. Few-shot fine-tuning classification results via 5-fold CV on Dataset 1. The performance metrics are summarized with the corresponding standard deviation (STD). For clarity, the top values are highlighted in bold black and the second highest in bold blue. "Param" represents parameters, and "FT" denotes traditional full fine-tuning.

| Method | Feature Fusion Stage | Finetuning Method | Param (M) | Tunable Param (M) | Accuracy | AUC |
|---|---|---|---|---|---|---|
| Image-wise CNN (Carneiro et al., 2017) | Late | FT | 44.7 | 44.7 | 0.765 ± 0.018 | 0.807 ± 0.016 |
| View-wise CNN (Wu et al., 2019) | Late | FT | 22.2 | 22.2 | 0.771 ± 0.023 | 0.813 ± 0.020 |
| Cross-view transformer (Tulder et al., 2021) | Middle | FT | 25.0 | 25.0 | 0.779 ± 0.013 | 0.817 ± 0.012 |
| Multi-view DeiT (Chen et al., 2022b) | Middle | FT | 5.5 | 5.5 | 0.777 ± 0.018 | 0.808 ± 0.025 |
| CLIP (Radford et al., 2021) | Early | FT | 151.3 | 151.3 | 0.545 ± 0.051 | 0.539 ± 0.081 |
| PointCLIP (Zhang et al., 2022) | Late | Image adapter | 151.3 | 1.2 | 0.629 ± 0.022 | 0.699 ± 0.022 |
| CLIP-Adapter (Gao et al., 2023) | Late | Image-text adapter | 151.3 | 0.6 | 0.606 ± 0.021 | 0.663 ± 0.028 |
| Mammo-CLIP (ViT-B/32) | Early | Image-text adapters | 151.3 | 1.3 | 0.785 ± 0.022 | 0.793 ± 0.025 |
| **Mammo-CLIP (ViT-B/16)** | Early | Image-text adapters | 151.3 | 1.3 | 0.790 ± 0.008 | **0.841 ± 0.017** |
| **Mammo-CLIP (ViT-L/14)** | Early | Image-text adapters | 363.8 | 4.1 | **0.823 ± 0.015** | 0.836 ± 0.019 |

## 4.2 Ablation study

**Impact of different mammographic views.** Table 3 examines the impact of different mammographic views on the classification performance across various Mammo-CLIP backbones. It yields two insights. First, when the two imaging views are used in isolation, the MLO view consistently outperforms the CC view across all evaluated metrics for all backbone architectures (ViT-B/32, ViT-B/16, ViT-L/14). Notably, the ViT-L/14 model on the MLO view demonstrates better performance (AUC = 0.827 ± 0.014, accuracy = 0.788 ± 0.018) compared to its CC view counterpart (AUC = 0.805 ± 0.012, accuracy = 0.761 ± 0.014). Second, fusing CC and MLO views significantly boosts classification performance compared to using each view separately. The ViT-L/14 model exhibits the highest performance in terms of accuracy (0.823 ± 0.015), while the ViT-B/16 model attains the highest AUC value (0.841 ± 0.017). This demonstrates the importance of combining information from both views to improve diagnostic outcomes in mammographic image analysis.

Table 3. Summary of the performance metrics for different Mammo-CLIP backbones across mammographic imaging views

| Backbones | Views | # Images | Accuracy | AUC |
|---|---|---|---|---|
| ViT-B/32 | CC + MLO | 4 | 0.785 ± 0.022 | 0.793 ± 0.025 |
| ViT-B/16 | CC + MLO | 4 | 0.790 ± 0.008 | **0.841 ± 0.017** |
| ViT-L/14 | CC + MLO | 4 | **0.823 ± 0.015** | 0.836 ± 0.019 |
| ViT-B/32 | CC | 2 | 0.714 ± 0.019 | 0.742 ± 0.024 |
| ViT-B/16 | CC | 2 | 0.740 ± 0.015 | 0.770 ± 0.036 |
| ViT-L/14 | CC | 2 | 0.761 ± 0.014 | 0.805 ± 0.012 |
| ViT-B/32 | MLO | 2 | 0.738 ± 0.028 | 0.774 ± 0.042 |
| ViT-B/16 | MLO | 2 | 0.761 ± 0.023 | 0.794 ± 0.030 |
| ViT-L/14 | MLO | 2 | 0.788 ± 0.018 | 0.827 ± 0.014 |

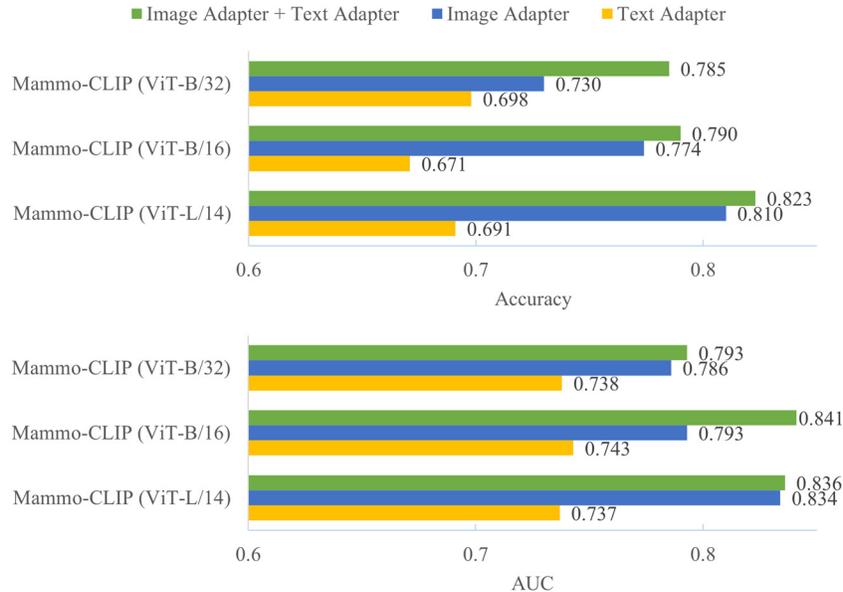

**Fig. 6.** Impact of image and text adapters on Mammo-CLIP performance.

**Impact of image and text adapters.** Fig. 6 compares the classification performance of Mammo-CLIP variants finetuned with image and/or text adapters. The joint use of both image and text adapters consistently yields superior performance across all tested model architectures. Notably, the Mammo-CLIP model with the ViT-L/14 backbone achieved the highest accuracy at 0.823 and an AUC score of 0.836, which is on par with the slightly higher AUC of 0.841 attained by the ViT-B/16 model. This near-equivalence in AUC, coupled with its superior accuracy, underscores the ViT-L/14 model's balanced strengths in classification performance, suggesting that its additional complexity indeed contributes to improved performance for our binary classification task. When comparing the individual use of image and text adapters, models finetuned with only image adapters generally outperform those fine-tuned with only text adapters in terms of accuracy and AUC. This trend suggests that the multimodal nature of the classification task derives more benefit from visual adaptation than from textual adaptation. However, the concurrent integration of both image and text adapters during the fine-tuning phase achieved the highest performance of the Mammo-CLIP models. This result highlights the value of multimodal learning approaches, illustrating their potential to enhance the models' efficacy in the domain-specific task of mammographic image classification.

**Table 4**. Summary of model-generated case classification accuracy (ACC) along with the standard deviation (STD) in 5-fold cross-validation on Dataset 1, with different numbers of local and global Transformer blocks (using Mammo-CLIP model with the ViT-B/32 backbone).

| # Local blocks | 0 | 2 | 4 | 6 | 8 | 10 | 12 |
| --- | --- | --- | --- | --- | --- | --- | --- |
| # Global blocks | 12 | 10 | 8 | 6 | 4 | 2 | 0 |
| Fold 1 | 0.795 | 0.805 | 0.800 | 0.742 | 0.747 | 0.721 | 0.737 |
| Fold 2 | 0.805 | 0.784 | 0.779 | 0.768 | 0.758 | 0.763 | 0.674 |
| Fold 3 | 0.805 | 0.774 | 0.758 | 0.716 | 0.747 | 0.705 | 0.674 |
| Fold 4 | 0.747 | 0.763 | 0.742 | 0.721 | 0.711 | 0.700 | 0.647 |
| Fold 5 | 0.772 | 0.772 | 0.751 | 0.735 | 0.720 | 0.688 | 0.693 |
| Mean ACC±STD | 0.785±0.022 | 0.780±0.014 | 0.766±0.021 | 0.736±0.018 | 0.737±0.018 | 0.715±0.026 | 0.685±0.030 |

**Impact of local vs. global transformer blocks.** We examined the influence of feature fusion location on classification accuracy by varying the configurations of local and global Transformer blocks. Table 4 summarizes the accuracy and standard deviation from 5-fold cross-validation applied to Dataset 1, employing the Mammo-CLIP model with a ViT-B/32 backbone. The data from the table shows a clear pattern: employing all transformer blocks for global feature fusion at an early stage results in the highest mean accuracy (0.785 ± 0.022). As the number of local blocks increases—and consequently, the number of global blocks decreases—there is a noted decline in performance. This trend suggests that leveraging the full complement of Transformer blocks for global fusion is particularly advantageous for multi-view mammogram analysis tasks.

**4.3 Zero-shot classification performance on Dataset 2**

Fig. 7 presents the ROC curves and AUC values for the zero-shot classification performance of 10 CAD models on Dataset 2. The results demonstrate that the Mammo-CLIP model implemented with the ViT-L/14 backbone achieves the highest AUC compared to other existing image-based models, with an AUC of 0.837 ± 0.034 (95% confidence intervals (CI) 0.778–0.886). In comparison, the best-performing benchmark model, the Cross-view transformer, attains an AUC of 0.807 ± 0.036 (95% CI 0.752–0.867]). Overall, Mammo-CLIP (ViT-L/14) outperforms all the benchmark models.

Fig. 8 displays the precision-recall curves of the zero-shot classification results for our model and the benchmark models. Consistent with the ROC analysis, Mammo-CLIP (ViT-L/14) achieves the highest PRAUC value compared to other methods. This shows that Mammo-CLIP (ViT-L/14) maintains a better balance between precision and recall across various threshold settings, suggesting its superior ability to correctly identify positive cases (malignant mammograms) while minimizing false positives (benign mammograms incorrectly classified as malignant) throughout the range of possible decision thresholds.

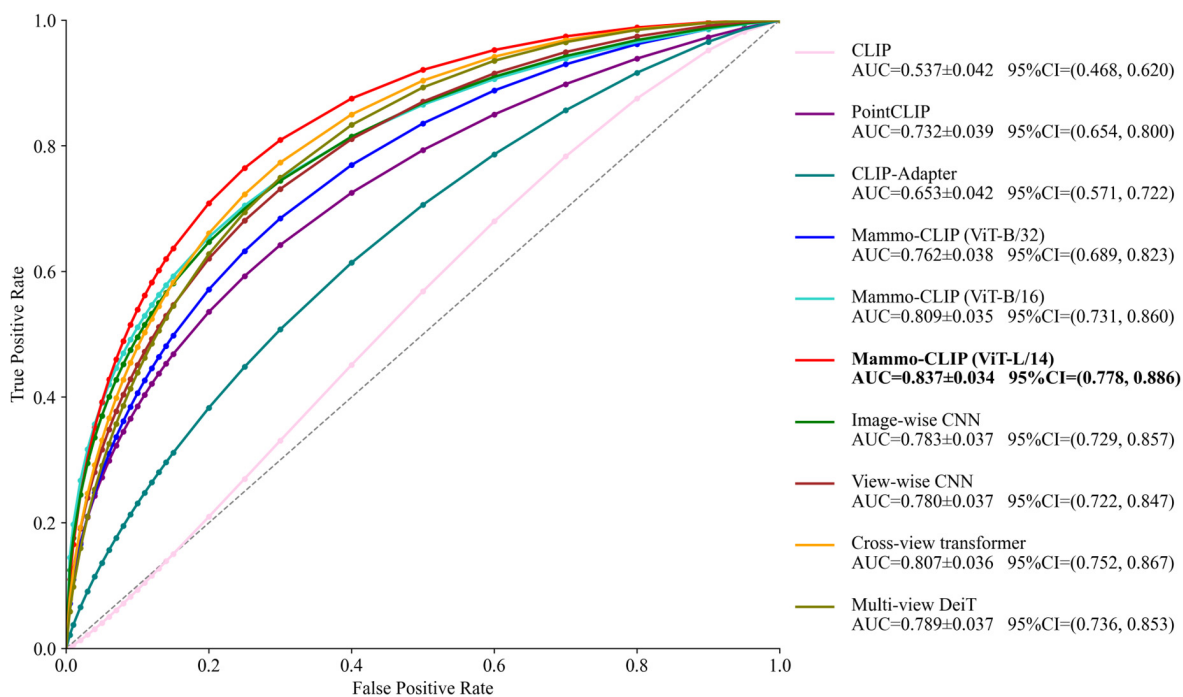

**Fig. 7.** ROC curves of zero-shot classification results on Dataset 2.

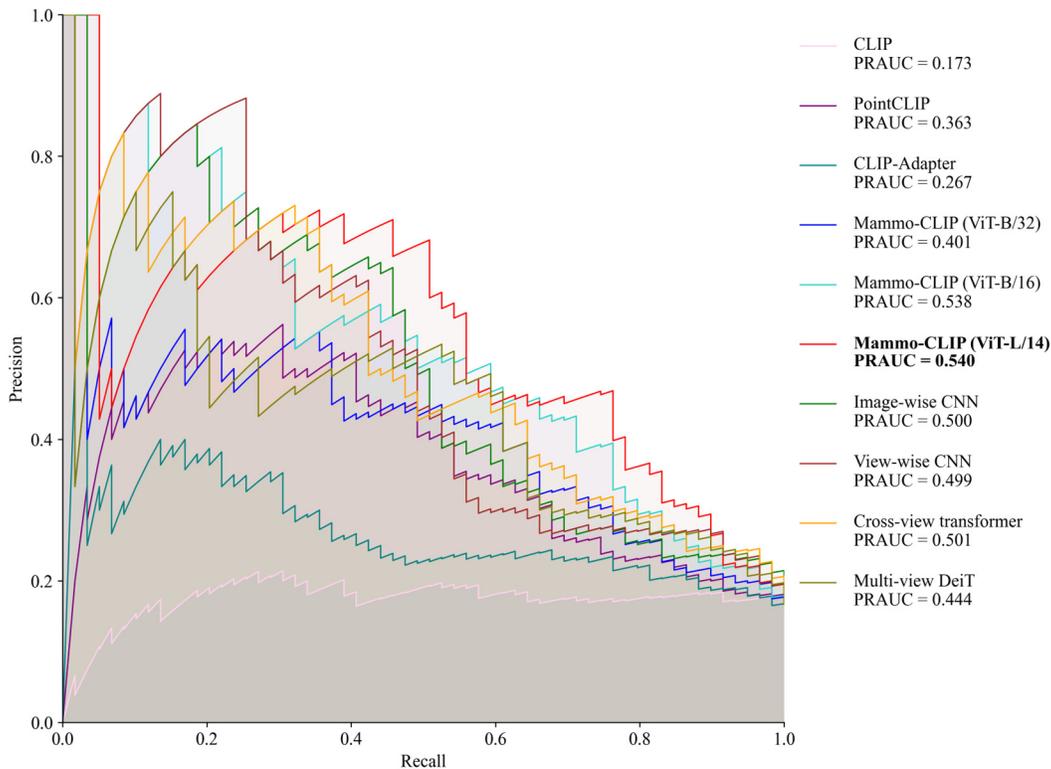

**Fig. 8.** Precision-recall curves of zero-shot classification results on Dataset 2.

## 5. Discussion

CAD schemes of multi-view mammograms have evolved dramatically over the last few decades, transitioning from traditional machine-learning approaches to advanced deep-learning technologies. Despite recent progress, most vision models in CAD development remain limited to single-modal training, relying exclusively on image data. Foundation VLMs, pre-trained with multimodal information (images and text), have demonstrated notable improvements in generalizability and have gained attention in medical imaging for tasks such as zero-shot and few-shot classification, detection, and generating medical reports, but their application in mammographic breast cancer screening has been limited (Zhao et al., 2023). To bridge this gap, we present Mammo-CLIP, an innovative framework for multi-view mammogram analysis derived from CLIP. Our framework establishes a multimodal CAD approach for breast cancer diagnosis, utilizing both multi-view mammograms and associated domain-specific text descriptions. Compared to single-modal approaches, Mammo-CLIP leverages the strengths of visual and textual data to enhance accuracy and generalizability in diagnosing breast cancer. To the best of our knowledge, Mammo-CLIP is the first VLM-based CAD scheme specifically tailored for multi-view mammogram analysis. This study distinguishes itself with several unique characteristics and contributes new insights to the field.

First, Mammo-CLIP significantly enhances multi-view mammogram analysis, addressing CLIP's limitation of handling only single-view images. Prior attempts to adapt CLIP for multi-view analysis introduced an inter-view adapter at the vision encoder's end, yielding unsatisfactory results—markedly inferior to traditional CNN-based methods. Our proposed multi-view mammogram analysis framework emphasizes the importance of performing multi-view feature fusion early within CLIP's vision encoder, as opposed to the commonly employed late-stage fusion approach. Early-stage fusion allows for the simultaneous use of low-level, mid-level, and high-level features, which is crucial for accurate breast cancer detection. This approach's effectiveness presumably stems from incorporating a broad range of diagnostic features, which better captures and leverages domain-specific knowledge, such as recognizing bilateral asymmetry patterns and tracking ipsilateral correspondence. These elements are crucial in enhancing CAD performance and significantly advancing its capabilities. In contrast, existing CAD approaches, whether CNN-based or hybrid transformer-based, typically execute multi-view feature fusion at mid or late stages. Our findings indicate that early-stage fusion is more advantageous, as shown in Table 4. By varying the number of local and global transformer blocks while maintaining their total count, we observed that an increase in global transformer blocks, which process features from all mammogram views concurrently, correlates with improved classification accuracy. This highlights the benefit of fusing multi-view mammogram features at the earliest stage using VLMs to develop high-performing CAD schemes. Interestingly, this strategy aligns with neuroscience insights on the early-stage natural integration mechanisms of diverse sensory information in the cortex for enhanced perception and decision-making (Macaluso, 2006).

Second, to investigate the influence of different mammographic views on classification performance within the Mammo-CLIP framework, we evaluated three backbone architectures for the vision encoder, each processing four images across CC and MLO views from both breasts and separately analyzing images from either view. As shown in Table 3, although both views offer valuable insights for malignancy detection, MLO-view images are particularly informative, resulting in superior model performance compared to CC-view images. In isolated view analyses, the MLO view surpasses the CC view in all metrics across the tested backbone architectures (ViT-B/32, ViT-B/16, ViT-L/14). Remarkably, the ViT-L/14 model analyzing MLO views achieves better outcomes (accuracy: 0.788±0.018, AUC: 0.827±0.014) than its CC view analysis (accuracy: 0.761±0.014, AUC: 0.805±0.012). The observed advantage of utilizing the MLO view compared to the CC view is consistent with previous research findings (Heidari et al., 2019).

Third, the integration of all four mammograms from both views significantly enhances Mammo-CLIP's classification capabilities by harnessing distinctive features that might be overlooked if each view were analyzed independently. This two-side-two-view approach achieves the highest performance, with an accuracy of 0.823±0.015 and an AUC score of 0.841±0.017. By imitating radiological practices, where information from the ipsilateral and bilateral mammograms is considered concurrently, this integrated approach suggests that effective image fusion boosts the CAD scheme's ability to distinguish between malignant and benign cases. Thus, our study not only affirms the potential of multi-view imaging-based CAD schemes for advancing future diagnostic technologies but also demonstrates the value of integrating such schemes within a multi-modal mammogram analysis framework. To our knowledge, the combination of multi-view mammographic imaging and a multi-modal CAD framework is studied for the first time in the literature.

Fourth, we implemented cost-efficient strategies for applying foundational VLMs to multi-view mammogram analysis by employing PETL techniques on our small-sized dataset. Traditional full finetuning methods are infeasible for few-shot finetuning scenarios where training samples are scarce, and the dataset size does not support thorough optimization of large foundational models. In such contexts, fully finetuning the Mammo-CLIP model yields the worst performance compared to other methods, as shown in Table 2. In contrast, inserting and finetuning plug-and-play adapter modules with 4.1 million tunable parameters—only about 1% of the total parameters—delivered the best performance in terms of accuracy and AUC.

Fifth, to maximize the benefits of using adapters, we introduced a novel adapter integration method that are different from the previous adapter insertion strategies. Instead of attaching a single adapter module outside the vision and text encoders (Gao et al., 2023; Zhang et al., 2022), we embedded adapters within each transformer block of both encoders. This design enhances the model's ability to analyze multi-view mammograms by adapting both image and textual feature representations at various levels to our specific task. A key innovation is the simultaneous fine-tuning of both visual and text adapters. Despite previous findings from CLIP-Adapter suggesting that fine-tuning the text adapter could interfere with image adapter performance (Gao et al., 2023), our approach demonstrates improvement rooted in the internal deployment of multiple adapters. By enabling thorough feature adaptation throughout the encoders, this new approach outperforms a single external adapter that only alters the final output.

Last, it is also worth noting that the adapters we employed represent the simplest form of PETL introduced in 2019. More sophisticated and efficient adapter variants like LoRA (Hu et al., 2021) have since emerged, promising enhanced adaptability of large-scale pre-trained models to specific tasks or domains. Incorporating these advanced adapter forms into our multi-view mammogram analysis framework could further enhance performance. Nevertheless, the superior results of Mammo-CLIP, even without the most cutting-edge adaptation techniques, highlight the efficacy of our new approach and its potential to lead in performance.

Despite the promising results, it is essential to acknowledge certain limitations and propose directions for future research. First, this study only focused on a case-based binary classification task that involves resizing mammograms from their initial high resolution to a 224×224 dimension. Although this resizing practice has been commonly used to fit the input dimensions of existing models (Carneiro et al., 2015, 2017), it may ruin fine-grained details discernible only at the original resolution, particularly for tasks like microcalcification detection, where the signs of malignancy are small and intricate. Previous research has also demonstrated that analyzing mammograms at their original resolution often yields the best results (Geras et al., 2017). Therefore, we recommend that future studies focus on adapting foundation VLMs to analyze high-resolution mammograms.

Second, the Mammo-CLIP framework utilizes relatively simple domain-specific textual descriptions for categorizing mammograms into benign or malignant classifications at the fine-tuning stage. These texts encapsulate essential information about mammographic views, breast laterality, and diagnostic labels to adapt the general-purpose CLIP model for the specialized task of analyzing multi-view mammograms (Fig. 3). While incorporating these basic textual descriptors into the text encoder has proven effective in boosting the image encoder's performance, we believe that leveraging more complex textual descriptions at the few-shot fine-tuning stage could offer even greater benefits. For instance, integrating specific information regarding a patient's history or symptoms alongside detailed descriptions of lesions could better exploit the synergy between visual and textual information. This, in turn, may lead to stronger zero-shot prediction capabilities.

## 6. Conclusion

As a prominent vision-language model, CLIP is gaining traction in various medical imaging tasks, yet its application in multi-view mammography analysis remains largely unexplored. In this study, we introduce a new Mammo-CLIP, which to the best of our knowledge is the first multi-modal framework designed to simultaneously process multi-view mammogram images and their associated domain-specific textual information. By incorporating early multi-view feature fusion and co-finetuning both visual and textual adapters, Mammo-CLIP obtains superior performance in malignancy classification compared to existing state-of-the-art multi-view image-based CAD schemes. These existing schemes are limited to single-modality image input, whereas Mammo-CLIP leverages both image and text data, resulting in improved accuracy and AUC on both internal and external validation datasets. This highlights the benefits of integrating multi-view mammogram images alongside domain-specific textual data to enhance the performance of CAD schemes in breast cancer detection or diagnosis. Our study results demonstrate Mammo-CLIP's potential as a strong and potentially superior alternative to single-modal CNNs and transformer-based models for building high-performing image-text-based CAD schemes tailored for multi-view digital mammograms. <u>We are preparing our code for public release to ensure its reproducibility and facilitate widespread use.</u> Further research is necessary to validate these results across more diverse datasets and explore the clinical implementation of this promising framework.

## CRediT Authorship Contribution Statement

**Xuxin Chen:** Conceptualization, Investigation, Methodology, Software, Writing - original draft, Visualization., **Yuheng Li:** Visualization, Writing - review & editing, **Mingzhe Hu:** Visualization, Writing - review & editing, **Ella Salari:** Visualization, Writing - review & editing, **Xiaoqian Chen:** Writing - review & editing, **Richard L.J. Qiu:** Visualization, Writing - review & editing, **Bin Zheng:** Data curation, Writing - review & editing, **Xiaofeng Yang:** Conceptualization, Resources, Writing - review & editing, Supervision, Funding acquisition.

## Declaration of Competing Interest

The authors declare that they have no known competing financial interests or personal relationships that could have appeared to influence the work reported in this paper.

## Declaration of Generative AI in Scientific Writing

During the preparation of this work the authors used ChatGPT 4 and Claude 3 Opus in order to improve language and readability. After using this tool/service, the authors reviewed and edited the content as needed and take full responsibility for the content of the publication.

## Acknowledgements

This research is supported in part by the National Institutes of Health under Award Number R56EB033332, R01EB032680, and R01CA272991.